\documentclass[10pt,twocolumn,letterpaper]{article}
\usepackage{authblk}
\usepackage{adjustbox}
\usepackage{algorithmic}
\usepackage[ruled,vlined]{algorithm2e}
\usepackage{amsmath,amssymb,amsfonts}
\usepackage{bm}
\usepackage{caption}
\usepackage{cite}
\usepackage{epsfig}
\usepackage{graphicx}
\usepackage{iccv}
\usepackage{times}
\usepackage{lscape}
\usepackage{lipsum,multicol}
\usepackage{rotating}
\usepackage{romannum}
\usepackage{subcaption}
\usepackage[pagebackref=true,breaklinks=true,colorlinks,bookmarks=false]{hyperref}

\DeclareMathOperator*{\argmax}{arg\,max}
\DeclareMathOperator*{\argmin}{arg\,min}

\usepackage[breaklinks=true,bookmarks=false]{hyperref}

\iccvfinalcopy 

\def\iccvPaperID{****} 

\ificcvfinal\fi

\begin{document}

\title{Extending Contrastive Learning to Unsupervised Coreset Selection}

\author[1]{Jeongwoo Ju}
\author[2]{Heechul Jung}
\author[2]{Yoonju Oh}
\author[1]{Junmo Kim}
\affil[1]{KAIST} \affil[2]{Kyungpook National University}

\maketitle

\begin{abstract}
Self-supervised contrastive learning offers a means of learning informative features from a pool of unlabeled data. 
In this paper, we delve into another useful approach---providing a way of selecting a core-set that is entirely unlabeled. In this regard, contrastive learning, one of a large number of self-supervised methods, was recently proposed and has consistently delivered the highest performance. This prompted us to choose two leading methods for contrastive learning: the simple framework for contrastive learning of visual representations (SimCLR) and the momentum contrastive (MoCo) learning framework. We calculated the cosine similarities for each example of an epoch for the entire duration of the contrastive learning process and subsequently accumulated the cosine-similarity values to obtain the coreset score. Our assumption was that an sample with low similarity would likely behave as a coreset. Compared with existing coreset selection methods with labels, our approach reduced the cost associated with human annotation. The unsupervised method implemented in this study for coreset selection obtained improved results over a randomly chosen subset, and were comparable to existing supervised coreset selection on various classification datasets (e.g., CIFAR, SVHN, and QMNIST).
\end{abstract}

\section{Introduction}
\label{sec:intro}

\begin{figure}[ht!]
\vspace{-2mm}
\begin{center}
\includegraphics[width=1\linewidth]{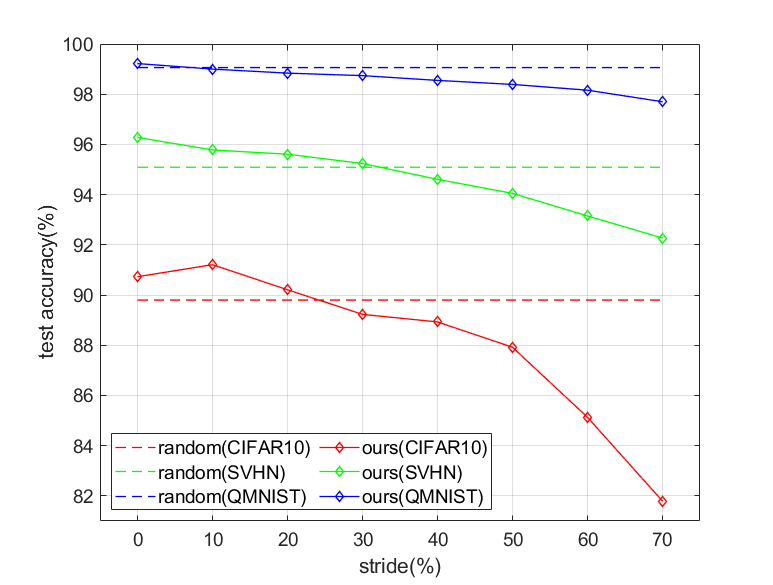}
\vspace{-7mm}
\end{center}
\caption{\textbf{Our coreset based on SimCLR performance on three datasets (CIFAR10, SVHN, QMNIST).} The horizontal line depicts the stride as a percentage of the training set size. The dotted and solid lines represent the test accuracy of the classification network trained with a randomly chosen subset and our coreset, respectively. The size of both subsets is 30\% of each dataset, i.e., 15,000, 22,500, and 18,000 samples for CIFAR10, SVHN, and QMNIST, respectively. At a stride of 0\%, our coreset achieves high performance over a random subset. As the stride moves to the right, the resulting accuracy decreases. This signifies that the corresponding examples exhibit a low coreset score.
We use ResNet18 architecture for both the coreset selection and classification.}
\label{fig1}
\vspace{-4mm}
\label{fig:observation}
\end{figure}

Deep learning-based methods have been highly effective in performing computer vision tasks such as image classification\cite{lee2018image}, object detection\cite{jiao2019survey}, and semantic segmentation\cite{girisha2019performance}. However, these methods generally require large amounts of data to produce accurate results; in particular, human annotation, an essential part of supervised learning, can be considerably time consuming and costly to implement. To address this problem, we selected a subset of the entire training set: given a training set ${D}$, we selected a subset ${D}^{'} \subset {D}$. This process requires identifying useful samples that can collectively represent the entire dataset. In this case, the underlying assumption is that not all samples contribute equally to a given task. In other words, each sample has a different level of contribution to the task---mostly referred to in the literature as high-contribution examples of the \textbf{coreset}. This assumption is a necessary improvement over a randomly selected subset (This topic has been thoroughly reviewed \cite{toneva2018empirical}, \cite{coleman2019selection}.). These results can be divided into two groups: supervised and unsupervised coresets (described in detail in Section \ref{sec:related}). 

One of the limitations of employing supervised coresets is that they do not reduce the annotation cost because they require a fully labeled dataset to execute successfully. Nonetheless, supervised coresets have been extensively studied, whereas unsupervised coresets have not received sufficient attention. Therefore, we aimed to select a coreset without human intervention, as is often the case when labels are not available. To the best of our knowledge, only Valvano et al. \cite{valvano2018unsupervised} have addressed the problem of unsupervised coreset selection. However, their method does not function desirably with large datasets. Moreover, their results were declared not robust; further research would be required to prove the usability of this method. 
Through this study, we aimed to perform unsupervised coreset selection using a deep learning approach.
It must be noted that a coreset provides more information about the parent dataset than a non-coreset does to complete the target task (e.g., classification). Therefore, it is reasonable to suppose that information-related metrics and target tasks are required to enable coreset selection.  
Because it is impractical to build a target task in the absence of a label, we adopted an alternative method of building a task, namely, self-supervised learning.
\begin{figure*}[t!]
\centering
\vspace{-3mm}
\includegraphics[width=\linewidth]{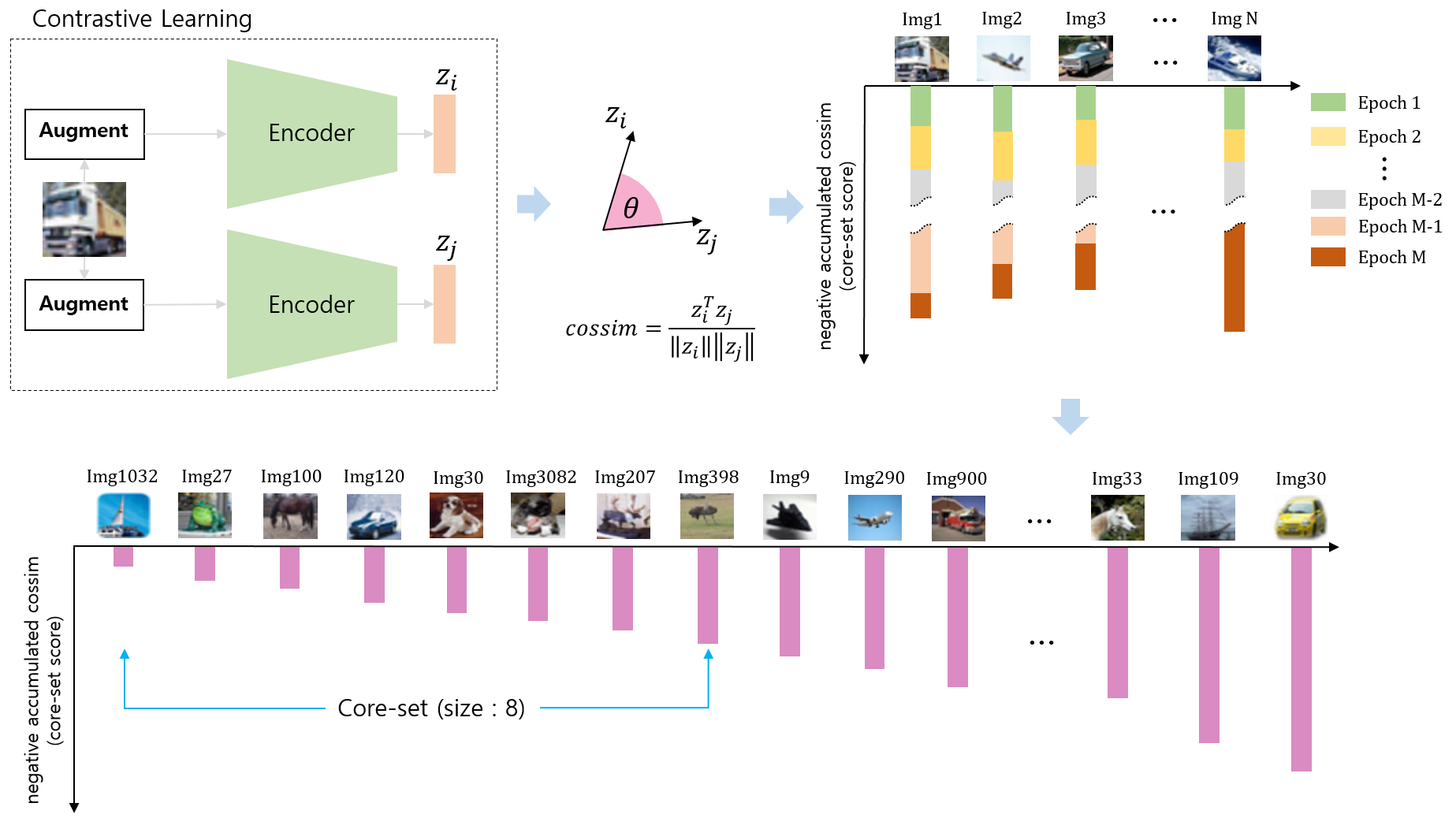}
\caption{\textbf{Overall flowchart of our coreset selection approach.} Example of our coreset selection: During contrastive learning (\textbf{top left}), the cossim value is calculated (\textbf{top middle}), and its negative value is accumulated (\textbf{top right}) simultaneously for each example. $N$ and $M$ represent the size of the training set and the number of epochs, respectively. After the learning is completed, the training data is sorted in increasing order of the coreset score (\textbf{bottom}). Lastly, the coreset with a specified size (e.g., eight) is selected. }
\vspace{-3mm}
\label{fig:overall}
\end{figure*}

Self-supervised learning has arisen for processing unlabeled data. With the objective of learning useful features from a pool of unlabeled data, self-supervised learning generally demands the description of a pretext task within the loss function; good performance depends on the definition of the pretext task. Hence, several researchers have attempted to devise a pretext task to enhance downstream task performance. However, it was recently reported that contrastive learning offers improved performance over existing pretext tasks. As its name implies, intermediate representations of the deep model are created to maximize the \textbf{cossim}(cosine similarity) between different views of the same example. 
As it is a state-of-the-art method, we decided to focus on contrastive learning in our study. Moreover, with respect to information-related metrics, we conjectured that the cossim would be a valid metric for coreset selection. This is because we assumed that the coreset would exhibit a low cossim value (in Section \ref{sec:ucs}).
As depicted in Fig. \ref{fig:overall}, we stored the negative accumulated cossim values for each example at the end of the epoch. We assigned the saved cossim value as the coreset score and sorted the examples according to their scores. Finally, we chose a final coreset with a specified size.
Subsequently, we compared the classification accuracy of the model trained from our coreset examples and from the randomly chosen examples separately by assigning equal sizes to the subsets. The empirical results showed that the accuracy of our coreset was more accurate than that of the random subsets (Section \ref{sec:exp}).
Notably, a low cossim(=high core-set score) value means that its examples is not easy for contrastive learning to predict compared with other examples with large cossim values.
Therefore, we inferred that the particular coreset has a characteristic that exhibits a low cossim value. This empirical evidence is in line with forgetting events \cite{toneva2018empirical}---an event where the sample is wrongly classified during training. In a previous study, it was discovered that an example with a large number of forgetting events had a high level of contribution in supervised learning. The opposite is also true, i.e., the lower the number of forgetting events, the smaller the contribution of the corresponding example).
Our work bridges the gap between contrastive learning and coreset selection and provides a new direction for further research. 
In this paper, we demonstrate the possibility of transferring an abstract space built from contrastive learning to identify coreset examples.
Our contributions to the literature are summarized as follows:
\begin{itemize}
\item To the best of our knowledge, our study is the first of its kind to substantiate that self-supervised learning is a highly suitable method for selecting a subset to generalize deep neural networks.
\item To measure the information of an example, we establish a coreset score as a function of cossim.
\end{itemize}
Our work is expected to provide guidance on the use of unsupervised coresets. Our code can be accessed from the following website: \footnote{\url{https://github.com/heechul-knu/self_supervised_learning_without_data}}.

\section{Related Work}
\label{sec:related}
In our study, we bridged the gap between contrastive learning and unsupervised coreset selection.
To establish context, we provide a brief history of self-supervision and revisit previous studies on coresets. 
\newline\newline
\noindent\textbf{Self-supervised learning} Self-supervised learning aims to define a pretext task to enable a model to learn useful features without the presence of labeled data. In this case, downstream tasks (classification, detection, etc.) serve as a tool to demonstrate the significance of the pretext task. In the last decade, numerous researchers have reported that their pretext tasks were more effective in downstream tasks. The major works are as follows: Context prediction \cite{doersch2015unsupervised} assigns a task to a model to predict the position of a patch relative to that of the center-cropped patch. Solving a jigsaw puzzle \cite{noroozi2016unsupervised} recasts a self-supervised problem by learning the permutation index of cropped patches. The counting task \cite{noroozi2017representation} is based on the concept that the sum of visual primitives from each patch must be equal to that of an entire image.

The above-mentioned studies were mainly based on the use of image patches. Unlike this stream, 
rotation \cite{gidaris2018unsupervised}, PIRL \cite{misra2019self}, MoCo v1 \cite{he2019momentum}, SimCLR \cite{chen2020simple}, and MoCo v2 \cite{chen2020improved} used an entire image as input.
The prediction angle at which an image is rotated is technically sounds and straightforward \cite{gidaris2018unsupervised}. Defining positive and negative pairs and maximizing the cossim value of positive pairs are key factors in contrastive learning, as mentioned in the MoCo v1-2 \cite{he2019momentum}\cite{chen2020improved} and SimCLR \cite{chen2020simple} papers. Both SimCLR and MoCo describe a positive pair as two augmented versions of the same example. SimCLR realizes contrastive learning via a multilayer perceptron (MLP) projection network, heavy data augmentation, and layer-wise adaptive rate scaling (LARS) optimizer \cite{you2017large}. On the other hand, MoCo mainly relies on a building dictionary equipped with a momentum encoder.
\newline
\newline
\noindent\textbf{Coreset Selection} A study on coresets was conducted with the goal of achieving data-processing efficiency with a given label (i.e., achieving faster learning and consuming lesser storage sapce)
Here, the term coreset refers to the most informative set of examples identified. To aid the reader’s understanding, we categorized existing studies into four groups depending on the feature type and the presence or absence of a label: \romannum{1}) hand-crafted feature + label: Bayesian inference \cite{campbell2018bayesian}, SVM \cite{tsang2005core}, Bayesian logistic regression \cite{huggins2016coresets}, submodular functions \cite{tschiatschek2014learning}\cite{wei2014submodular}\cite{wei2015submodularity}, k-means and k-median clustering \cite{har2004coresets}\cite{har2007smaller}, Gaussian mixture models \cite{lucic2017training}, optimization framework \cite{ghadikolaei2019learning}, \romannum{2}) hand-crafted feature + no label: Submodular optimization for speech data\cite{ni2015unsupervised}, \romannum{3}) learnable feature + label: forgetting events \cite{toneva2018empirical}, selection via proxy(SVP) \cite{coleman2019selection}. \romannum{4}) learnable feature + no label: Unsupervised data selection \cite{valvano2018unsupervised}---our work falls into this category.

In particular, a few attempts focusing on the application of deep neural networks (DNNs) were made until Toneva et al.\cite{toneva2018empirical} discovered a correlation between coresets and forgetting events. Furthermore, Coleman et al. \cite{coleman2019selection} extended this work by employing a proxy model for selection. Here, the target model was not necessarily identical to the model used for selecting the coreset, but a smaller network could achieve this selection. Valvano et al. \cite{valvano2018unsupervised} considered an unsupervised coreset selection; their work was based on a convolutional variational autoencoder (CVAE) to develop an embedding space in which the distance between successive features was calculated. Their work was limited to a small amount of data; furthermore, the computation was intractable, as a large amount of data is presented. 

\section{Unsupervised Coreset Selection}
\label{sec:ucs}

The objective of unsupervised coreset selection is to identify coreset examples in the absence of labels. Here, we recalled that a coreset provides more information than a non-coreset. In other words, an information-related metric must be established to measure the coreset score. Because contrastive learning is the fundamental idea of our work, the cossim can be readily obtained. Therefore, we were able to prove that cossim is a valid metric to represent the coreset. Our hypothesis is that cossim plays a major role in measuring the coreset score. This hypothesis was made based on visual inspection, which is presented in Section \ref{subsec:visualinspec}. First, to confirm our hypothesis, we present our observations.
\newline

\begin{figure}[t!]
\begin{center}
\vspace{-5mm}
\includegraphics[width=1\linewidth]{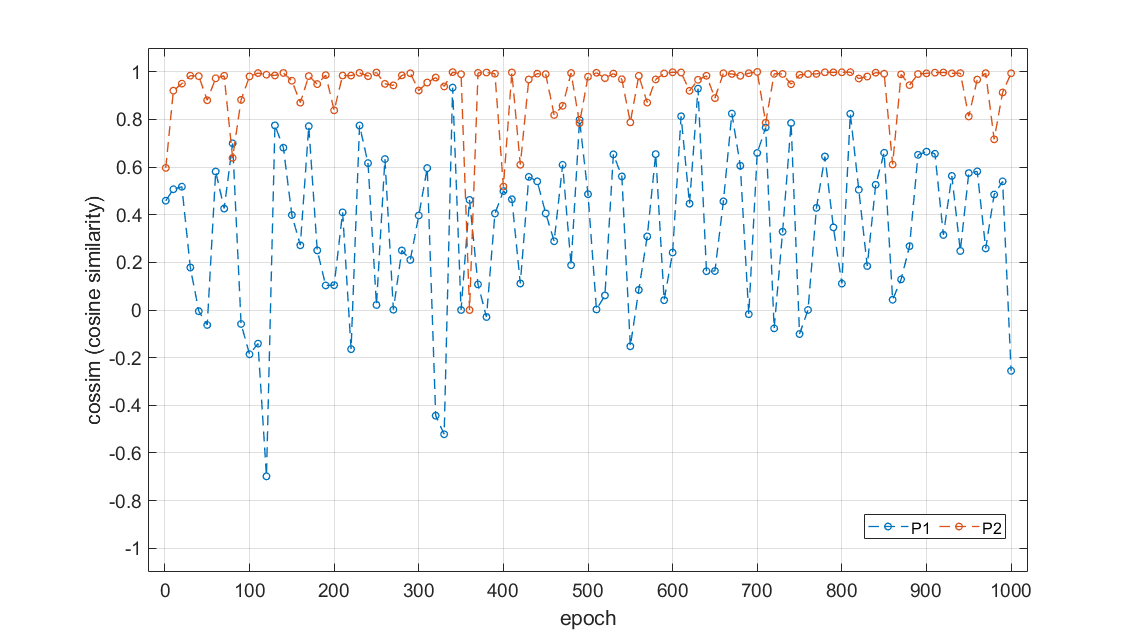}
\vspace{-5mm}
\end{center}
\caption{\textbf{Cossim values for each epoch.} Cossim of two examples (CIFAR10) with the highest and lowest coreset scores based on SimCLR: P1(\textbf{blue line}) and P2(\textbf{red line}). Note that P1 barely approaches 1 for the entire duration of the learning process. } 
\label{fig:tracking}
\vspace{-3mm}
\end{figure}

\noindent\textbf{Observations.} During contrastive learning, we calculated the coreset score as described in Alg. \ref{alg:score}.
The resulting array $A$ corresponds to a list of index examples in the training set. 
Provided that our score represents the coreset adequately, the performance increases according to the coreset score.
To provide a more concrete proof, we built a coreset score based on SimCLR and trained a classification network with cross-entropy loss and a subset of the training set that is described in Eq. \ref{eq:observation}. The stride $s$ ranges from 0\% to 70\% of the size of the training set and the subset size $L$ is set to 30\% of the size of the training set. $\bm{x}_k, \bm{y}_k$ represents a pair of $k$-th examples and targets, and $\ell(\cdot)$ is the cross-entropy loss. For example, in the case of CIFAR10, the sizes of the training and testing set are 50,000 and 10,000, respectively. We chose 30\% of the training set (15,000 examples) according to the coreset score and stride. In addition, we set the stride $s$ as follows: $s \in \{ 0, 5000 ,10000, 15000, 20000, 25000, 30000, 35000\}$. Following this, we trained the classification network with these subset samples and evaluated the performance with the testing set. 

\begin{equation}
Loss = \frac{1}{L} \sum_{k} \ell(\bm{x}_k, \bm{y}_k), \;\;\; k \in A[s : s+L ]. 
\label{eq:observation}
\end{equation}

Fig. \ref{fig:observation} illustrates the performance trend for the three datasets (CIFAR10, SVHN, and QMNIST). A clear correlation can be observed between the test accuracy and stride. As the stride increases (and the coreset score decreases), we observed that the performance of our coreset tends to degrade almost linearly across the datasets, and at approximately stride=20\%, it descends below that of the randomly generated subset. This observation emphasized the fact that the examples that provided more information compared with the others exhibited high coreset scores, and vice versa (the examples that were not too informative had a low coreset score). Additionally, we observed that cossim adequately represents the information-related metric required to measure the score. Hence, it was inferred that our score effectively sorted the examples according to the coreset. On the contrary, our coreset performance would exhibit a uniform trend across strides in a failure case.
\newline

\begin{algorithm}[b!]
\caption{\label{alg:score} Identify coreset}.
\begin{algorithmic}
\small
\STATE \hspace{-3mm} initialize metric $M[k]=0, k \in [n] $
\STATE \hspace{-3mm} \textbf{input:} subset size $L$
\STATE \textcolor{blue}{\# calculate the coreset score}
\STATE \hspace{-3mm} \textbf{while} not constrastive learning done \textbf{do}
\STATE \textbf{for all} $k \in [n]$ \textbf{do}
\STATE $~~~~$ $M[k]=M[k]-cossim(z_{i}(k), z_{j}(k))$ 
\STATE \textcolor{blue}{$~~~~$\# $z_{i}, z_{j}$ are a positive pair of latent variables for the k-th example}
\STATE \hspace{-3mm} $A=$argsort($M$, ascending order) 
\STATE \hspace{-3mm} \textbf{return} $A[0:L]$ $~~$\textcolor{blue}{\# resulting coreset}
\end{algorithmic}
\end{algorithm}

\noindent\textbf{Tracking cossim.} It is worth noting that averaging cossim (cossim accumulation) rather than using a single cossim is an efficient strategy. 
After SimCLR learning on CIFAR10 was completed, we selected two examples whose mean cossim were the highest and lowest, respectively, i.e., $x_{k'}, x_{l'},$ where $ k'=\argmax_{k} M[k]=A[0], \; l'=\argmin_{k} M[k]=A[N]$. Here, we tracked their cossim values for each epoch during learning, as shown in Fig. \ref{fig:tracking}, where P1 and P2 correspond to $x_{l'}, x_{k'}$, respectively. Unexpectedly, the P2 example nearly maintained its value close to 1 from the beginning of the learning. However, the value of P1 hardly approaches 1 even at the end of learning. In addition, the cossim values of both the examples abruptly changed over the entire learning process. 
Thus, considering noise-like cossim, averaging rather than choosing a single cossim value at a certain epoch in the middle of learning was a reasonable approach.
\newline

\begin{figure}[t!]
    \centering
    \begin{subfigure}[t]{0.22\textwidth}
        \centering
        \includegraphics[width=\textwidth]{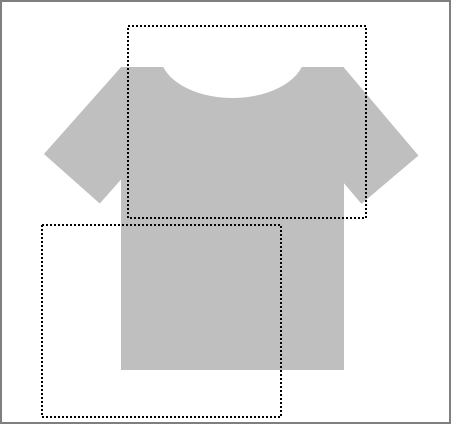}
        \caption{normal t-shirt}
        \label{fig:normal_tshirt}
    \end{subfigure}
    \hfill
    \begin{subfigure}[t]{0.22\textwidth}
        \centering
        \includegraphics[width=\textwidth]{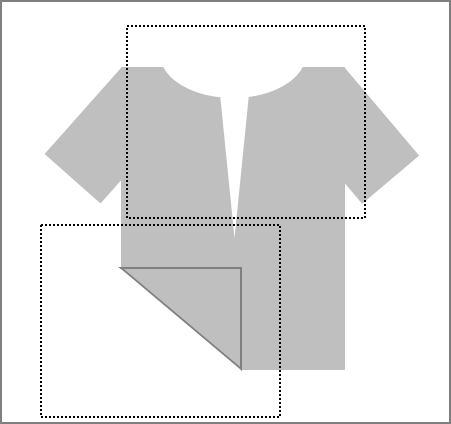}
        \caption{abnormal t-shirt}
        \label{fig:abnormal_tshirt}
    \end{subfigure}
    \caption{\textbf{Example of a normal and abnormal T-shirt.} The dashes represent two randomly cropped regions. In normal T-shirts, the contents of the two cropped regions are not significantly different. On the other hand, in the abnormal T-shirt, the content of the two cropped regions is significantly different.}
    \vspace{-3mm}
    \label{fig:tshirt}
\end{figure}

\begin{figure*}[ht]
    \vspace{-5mm}
     \centering
     \begin{subfigure}[t]{0.33\textwidth}
         \centering
         \includegraphics[width=\textwidth]{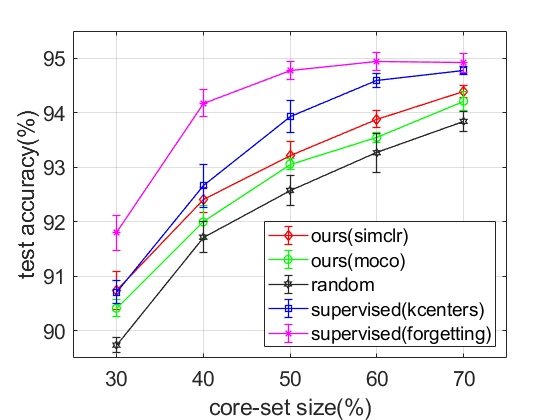}
         \caption{CIFAR10}
         \label{fig:fmnist_coreset}
     \end{subfigure}
     \hfill
     \begin{subfigure}[t]{0.33\textwidth}
         \centering
         \includegraphics[width=\textwidth]{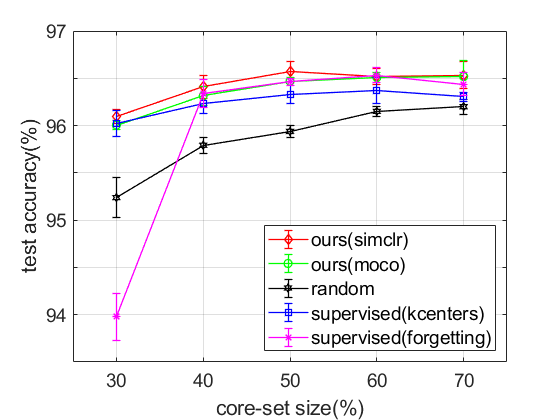}
         \caption{SVHN}
         \label{fig:svhn_coreset}
     \end{subfigure}
     \hfill
     \begin{subfigure}[t]{0.33\textwidth}
         \centering
         \includegraphics[width=\textwidth]{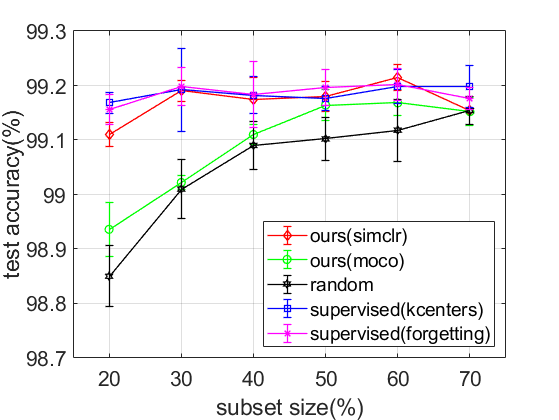}
         \caption{QMNIST}
         \label{fig:cifar_coreset}
     \end{subfigure}
        \caption{\textbf{Coreset selection performance on three classification datasets.} The vertical bars indicate the standard deviation of the accuracy. In most cases, our coreset was more accurate than the random subset. Despite the absence of labels, our results were comparable to those of supervised coreset selection (SCS). Notably, SCS (with forgetting events) was less accurate than other methods in the case of SVHN. 
        For QMNIST, we additionally examined the performance at 20\% subset size to verify the saturation point; above 20\% subset size, our SimCLR model and SCS had become saturated at top performance, whereas our MoCo model and random subset were still undergoing improvement.}
        \label{fig:coreset_acc}
\end{figure*}

\begin{figure*}[ht]
     \centering
     \begin{subfigure}[t]{0.33\textwidth}
         \centering
         \includegraphics[width=\textwidth]{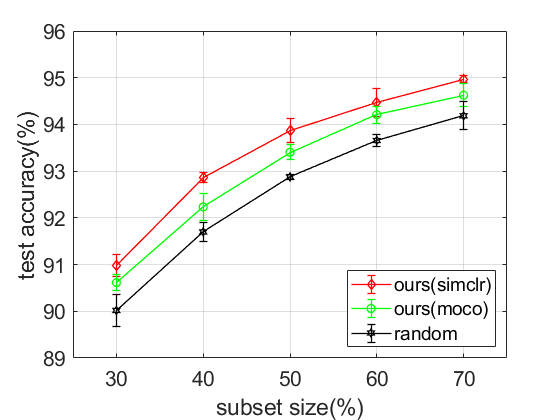}
         \caption{CIFAR10}
         \label{fig:agnostic_cifar10}
     \end{subfigure}
     \hfill
     \begin{subfigure}[t]{0.33\textwidth}
         \centering
         \includegraphics[width=\textwidth]{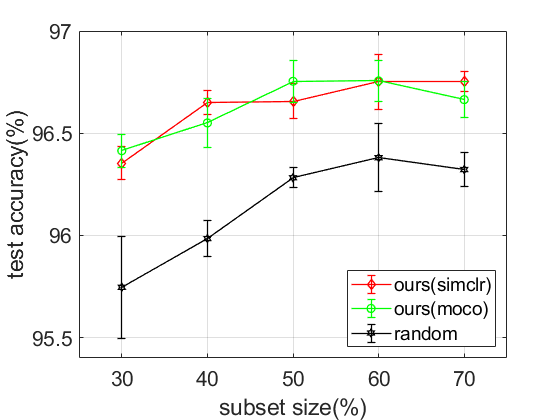}
         \caption{SVHN}
         \label{fig:agnostic_svhn}
     \end{subfigure}
     \hfill
     \begin{subfigure}[t]{0.33\textwidth}
         \centering
         \includegraphics[width=\textwidth]{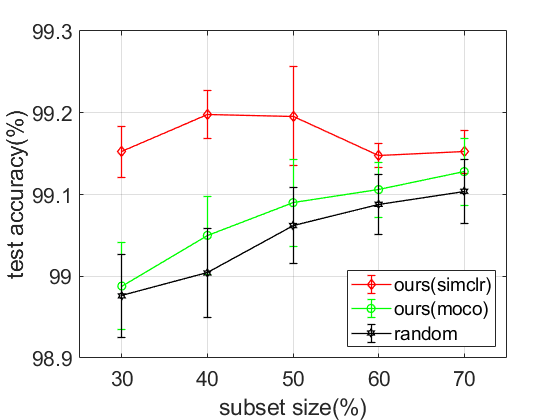}
         \caption{QMNIST}
         \label{fig:agnostic_qmnist}
     \end{subfigure}
        \caption{\textbf{Model-agnostic performance on three classification datasets.} Coreset selection model: ResNet18; target model: ResNet101. The vertical bars indicate the standard deviation of the accuracy. For all three datasets, our coreset was more accurate than the random subset.}
        \label{fig:agnostic}
        \vspace{-3mm}
\end{figure*}

\noindent\textbf{Identifying the coreset.} As mentioned above, we calculated the coreset score by averaging the cossim values. This allowed us to select the coreset based on the score. Specifically, our coreset score was calculated as follows: throughout the contrastive-learning process, the negative cossim value is accumulated for each example. Following this, we sorted the scores in ascending order. The final coreset was then chosen with the size of the value specified according to the score, as displayed in Alg.\ref{alg:score}, where $[n]=\{1,...,N\}$ and $N$ is the size of training set. This approach, although simple, has proven effective in identifying the coreset, as demonstrated in Section \ref{subsec:coreset}. It owes to the fact that in contrastive learning, strong augmentation plays a crucial role in realizing representation learning. For instance, considering two cases in which random cropping---a form of augmentation---is applied to two different types of T-shirts (Fig. \ref{fig:tshirt}), we can assume that the two randomly cropped regions include dissimilar contents, although they are in an identical class. Consequently, the network experiences difficulties in minimizing the value of cossim for abnormal T-shirts, resulting in lower coreset scores compared with those of normal T-shirts. Thus, we concluded that the abnormal T-shirt possibly provides more information about the concept of the T-shirt than the normal one did. We verify this claim in Section \ref{subsec:coreset}. 
Intuitively, the most informative example would include the maximum amount of content possible. The bottom line of this claim is that an example belonging to a coreset would exhibit an abnormal or peculiar appearance. 
\newline

\noindent\textbf{Assessing the coreset.} Any approach to obtain a reliable coreset must meet the following requirements: \romannum{1}) consistency: across random seeds for the selection model, the ranking of the resulting coreset score must be as stable as possible. \romannum{2}) model-agnostic: once the coreset is chosen, it should outperform a randomly chosen set regardless of the model of the target task. We set up an evaluation protocol and describe the results in Sections \ref{subsec:consistent} and \ref{subsec:model-agnostic}. Furthermore, the absence of labels connotes that the data imbalance among the interclasses is an inherent problem for unsupervised coreset selection. We also examine the class imbalance of our coreset compared to a randomly chosen coreset in Section \ref{subsec:imbalance}. 


\section{Experiments}
\label{sec:exp}
In this section, we prove that coreset selection is a valid approach for performing classification tasks. Our coreset is an improvement over the random performance of three different datasets (CIFAR10, SVHN, and QMNIST). For comparison with supervised coreset selection (SCS), we conducted SCS experiments, i.e., using greedy k-centers \cite{sener2017active} and forgetting events \cite{toneva2018empirical}, which are also referred to in the literature \cite{coleman2019selection}.
In most cases, the performance of SCS, which is label-hungry by nature, manifestly surpasses that of our method. However, SCS is arguably an invalid approach for cost reduction because it requires fully annotated data. The importance of our study is thus highlighted: our proposed method is designed to directly process unlabeled data although it is outperformed by SCS.
In addition, as mentioned in Section \ref{sec:ucs}, along with Fig. \ref{fig:tshirt}, we presented another experiment, the coreset cross test, designed to provide evidence that a set of examples chosen by our coreset score offer highly informative content.

\subsection{Implementation Details}
\label{subsec:imple}
\textbf{Datasets.} To vary the example type to gray-scale, RGB, digits, and objects, we chose three different datasets (CIFAR10\cite{krizhevsky2009learning}, SVHN\cite{netzer2011reading}, and QMNIST \cite{qmnist-2019}), which consisted of 10 classes each. 
CIFAR10 is intended to support object classification and consists of 50,000 training and 10,000 testing examples. Each example has a size of 32$\times$32; the examples include center-cropped objects and RGB channels. QMNIST datasets are generated digits based on NIST Special Database 19 \cite{grother1995nist} with the purpose of maintaining MNIST \cite{lecun1998gradient} preprocessing. 
QMNIST is divided into a training dataset of 60,000 images and a testing dataset of 10,000 images. These are gray-scale images with a resolution of 28$\times$28. In contrast to QMNIST, SVHN comprises RGB channels. We used 73,257 examples for training and 26,032 for testing. These are cropped images of digits with a size of 32$\times$32. We omitted the extra images from this dataset (531,131 examples).  For both SVHN and QMNIST, the annotated labels directly correspond to digits, i.e., label `1' is the digit 1, etc. 
\newline
\newline
\textbf{Models.} We deployed ResNet18 \cite{he2016deep} for the contrastive and classification tasks.
In SimCLR learning, our coreset was constructed on the basis of the code available on the website\footnote{\url{https://github.com/Spijkervet/SimCLR}}. We set the batch size to 1024, number of epochs to 1,000, dimension of projection to 128, and employed the LARS optimizer \cite{you2017large} across the datasets.
In MoCo learning, we shared all the hyperparameters, such as the epochs (600) and batch size (512), across the datasets, and used cosine learning--rate scheduling based on the code available on the website\footnote{\url{https://github.com/facebookresearch/moco}}
For both types of contrastive learning, augmentation techniques including randomly resized crops, random horizontal flip, color jittering, and random gray-scale were implemented. In particular, because QMNIST consists of gray-scale images, we modified the first layer of ResNet18 to contain only a single channel. Additionally, for both SimCLR and MoCo, we deployed the following augmentation techniques across datasets: randomly resized crop, color jittering, horizontal flip, and applying gray-scale.

\subsection{Coreset Selection Task}
\label{subsec:coreset}
For training SCS with forgetting events and greedy k-centers, we utilized an entirely labeled training set and selected a subset according to their coreset scores. For SVHN, whose training set size is 73257, we set subset sizes (30\% $\sim$ 70\%) as follows: $\{ 22500,~30000,~37500,~45000,~52500 \}$.
The results of our classification task are shown in Fig. \ref{fig:coreset_acc}. We calculated the mean and standard deviation of the test accuracy over five runs. Our coreset proved to exhibit a higher accuracy than the randomly chosen coreset did. This implies that in contrastive learning, cossim serves as a useful metric for selecting the coreset. In most cases, SCS outperforms our coreset, whereas for the SVHN dataset, our results are comparable despite being unlabeled. 
As QMNIST contained gray-scale images, and the dataset is relatively easy to manipulate compared with the others, our improvement was marginal; however, it outperformed the random coreset at all subset sizes. 
Across datasets, our performance lay between SCS and the random subset, highlighting the importance of our study; despite having no labels, we were able to increase performance close to that of SCS.

\subsection{Coreset Cross Test}
\label{subsec:coreset}
To verify whether the examples chosen by our coreset score are highly informative, we designed a cross test as follows: once the coreset score based on SimCLR was established, we split the original training set into two subsets; the top 30\% $\sim$ 70\% as the training set and the bottom 30\% as the testing set (referred to as coreset $\rightarrow$ non-coreset or C $\rightarrow$ N). Similarly, we set the bottom 30\% $\sim$ 70\% as the training set and the top 30\% as the testing set (referred to as the non-coreset $\rightarrow$ coreset or N $\rightarrow$ C), and repeated each test for five runs. As shown in Table. \ref{tb:cross_test}, in all the cases, the C $\rightarrow$ N cross test was superior to N $\rightarrow$ C. This result supports our claim that samples with high coreset scores provide more information than those with low scores. Furthermore, their appearances are significantly different (elaborated in Section \ref{subsec:visualinspec}).

\begin{table}[h!]
\caption{\textbf{Cross-test results.} Average ($\pm$ std) obtained from the five runs on the three datasets. C and N represent the coreset and non-coreset, respectively. Across the dataset and fraction of the subset, the C $\rightarrow$ N test achieved significantly higher accuracy than that of the N $\rightarrow$ C test. Cross tests were conducted using the coreset score based on SimCLR.}
\label{tb:cross_test}
\centering
\resizebox{\linewidth}{!}{
\begin{tabular}{ccccccc}
\hline\hline
         &                      &  \multicolumn{5}{c}{ Fraction of subset}  \tabularnewline 
 Dataset &        cross         &    30\%   &    40\%   &    50\%   &    60\%   &  70\% \tabularnewline \hline\hline
 \vspace{-1.1mm}
         &   C $\rightarrow$ N  &   94.27   &   95.44   &   96.14   &   96.68   &  96.99 \tabularnewline 
         &                      & $_{(0.14)}$ & $_{(0.25)}$ & $_{(0.08)}$ & $_{(0.16)}$ & $_{(0.04)}$ \tabularnewline
\vspace{-1.1mm}
 CIFAR10 &   N $\rightarrow$ C  &    67.39    &  73.95   & 80.18  & 83.54    &  86.68  \tabularnewline 
         &                      & $_{(0.80)}$ & $_{(0.35)}$ & $_{(0.49)}$ & $_{(0.24)}$ & $_{(0.14)}$ \tabularnewline \hline
\vspace{-1.1mm}
         &   C $\rightarrow$ N  &   98.24   &   98.38   &   98.50   &   98.50   &  98.80 \tabularnewline 
         &                      & $_{(0.09)}$ & $_{(0.04)}$ & $_{(0.04)}$ & $_{(0.06)}$ & $_{(0.06)}$ \tabularnewline
\vspace{-1.1mm}
 SVHN    &   N $\rightarrow$ C  &    83.30    &  85.48   & 86.90  & 88.31    &  89.76  \tabularnewline 
         &                      & $_{(0.25)}$ & $_{(0.41)}$ & $_{(0.13)}$ & $_{(0.15)}$ & $_{(0.24)}$ \tabularnewline \hline
\vspace{-1.1mm}
         &   C $\rightarrow$ N  &   99.88   &   99.91   &   99.92   &   99.91   &  99.92 \tabularnewline 
         &                      & $_{(0.02)}$ & $_{(0.01)}$ & $_{(0.01)}$ & $_{(0.02)}$ & $_{(0.01)}$ \tabularnewline
\vspace{-1.1mm}
 QMNIST  &   N $\rightarrow$ C  &    93.07   &  94.07   & 94.84  & 95.44    &  96.03  \tabularnewline 
         &                      & $_{(0.44)}$ & $_{(0.12)}$ & $_{(0.11)}$ & $_{(0.11)}$ & $_{(0.19)}$ \tabularnewline \hline \hline
\end{tabular}}
\end{table}

\begin{figure*}[ht!]
\vspace{-3mm}
    \begin{tabular}{cccc}
        \begin{subfigure}{5mm}{} \end{subfigure}
        \begin{subfigure}{55mm}\centering{~~~~CIFAR10} \end{subfigure}
        \begin{subfigure}{55mm}\centering{~~~~SVHN} \end{subfigure}
        \begin{subfigure}{55mm}\centering{~~~~QMNIST} \end{subfigure}
        \\
        \begin{subfigure}{5mm}\rotatebox{90}{\centering{~~~subset size : 30\%}} \end{subfigure}
        \begin{subfigure}{55mm}
            \includegraphics[width=55mm]{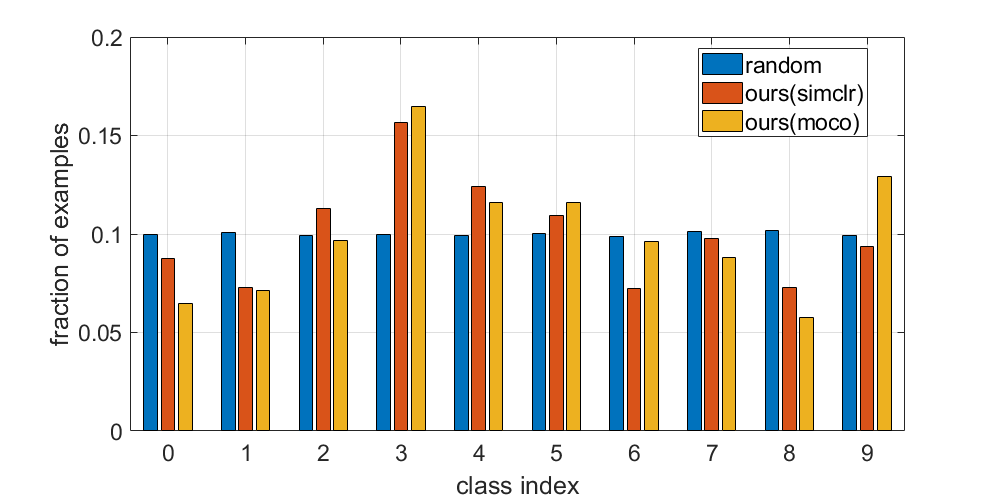}
            \caption{}
        \end{subfigure}
        \begin{subfigure}{55mm}
            \includegraphics[width=55mm]{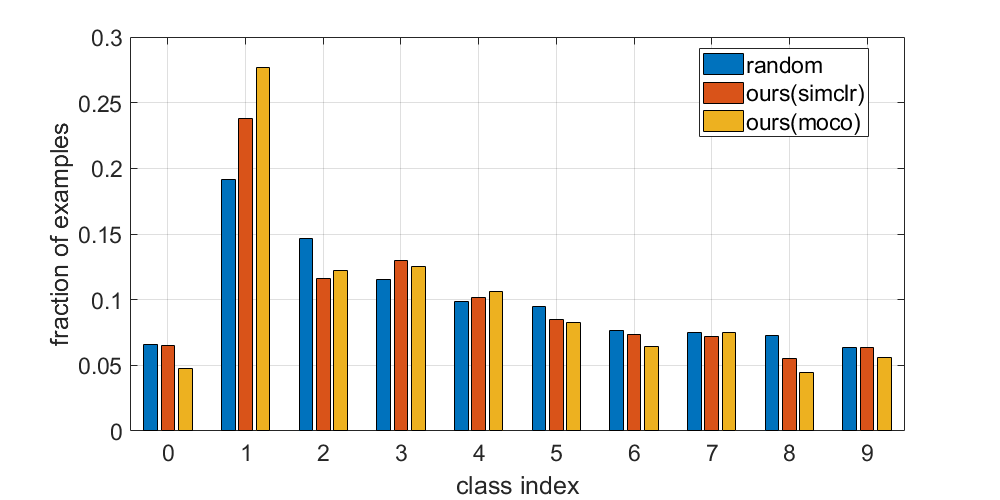}
            \caption{}
        \end{subfigure}
        \begin{subfigure}{55mm}
            \includegraphics[width=55mm]{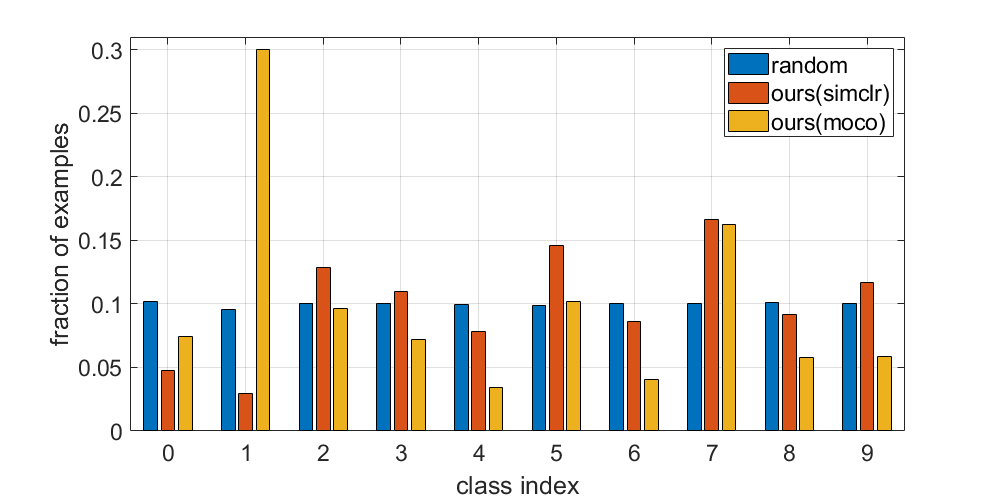}
            \caption{}
        \end{subfigure}
        \\
        \begin{subfigure}{5mm}\rotatebox{90}{\centering{~~~subset size : 50\%}} \end{subfigure}
        \begin{subfigure}{55mm}
            \includegraphics[width=55mm]{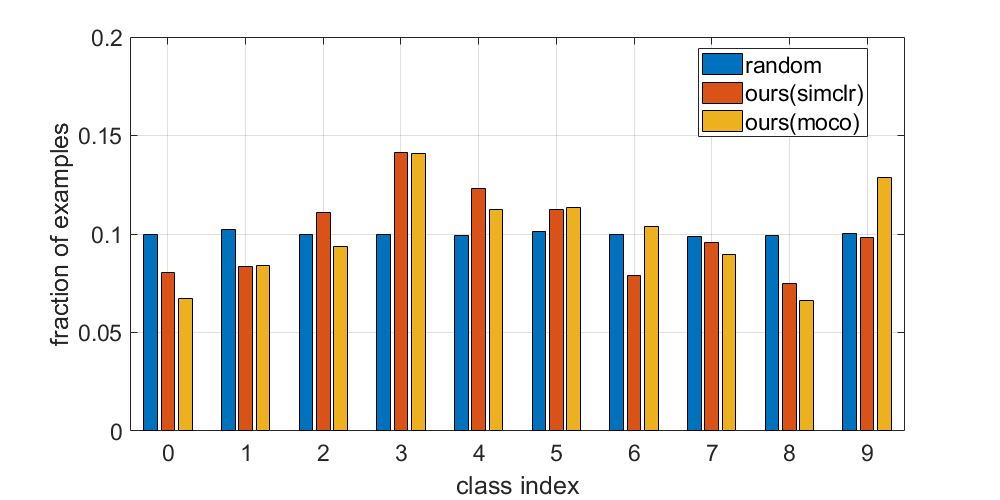}
            \caption{}
        \end{subfigure}
        \begin{subfigure}{55mm}
            \includegraphics[width=55mm]{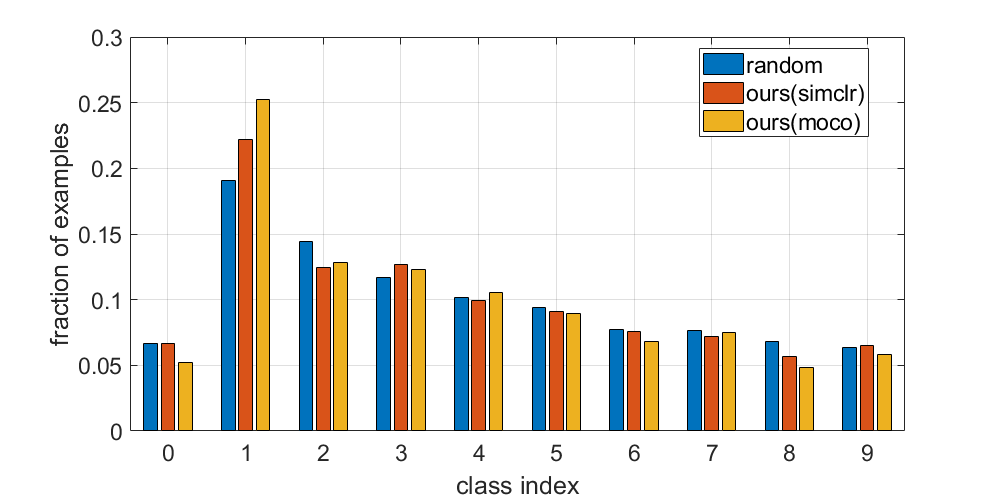}
            \caption{}
        \end{subfigure}
        \begin{subfigure}{55mm}
            \includegraphics[width=55mm]{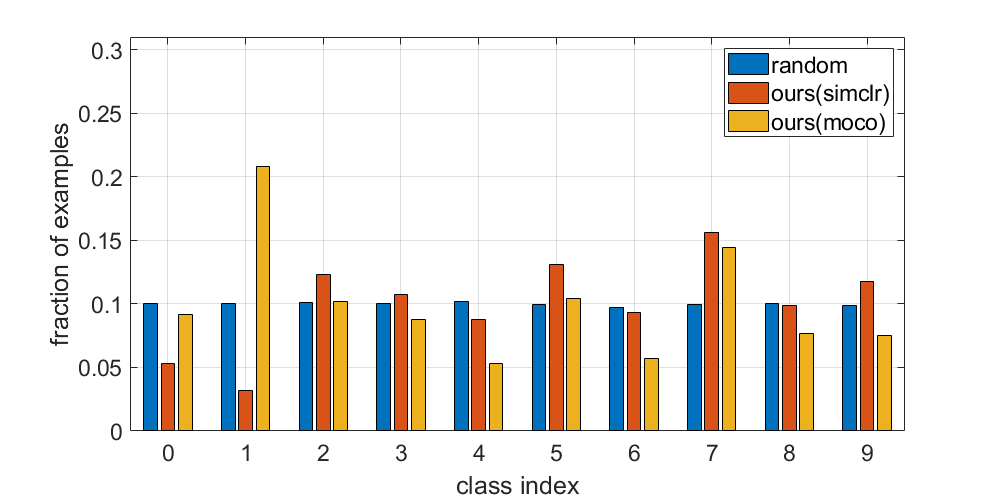}
            \caption{}
        \end{subfigure}
        \\
        \begin{subfigure}{5mm}\rotatebox{90}{\centering{~~~subset size : 70\%}} \end{subfigure}
        \begin{subfigure}{55mm}
            \includegraphics[width=55mm]{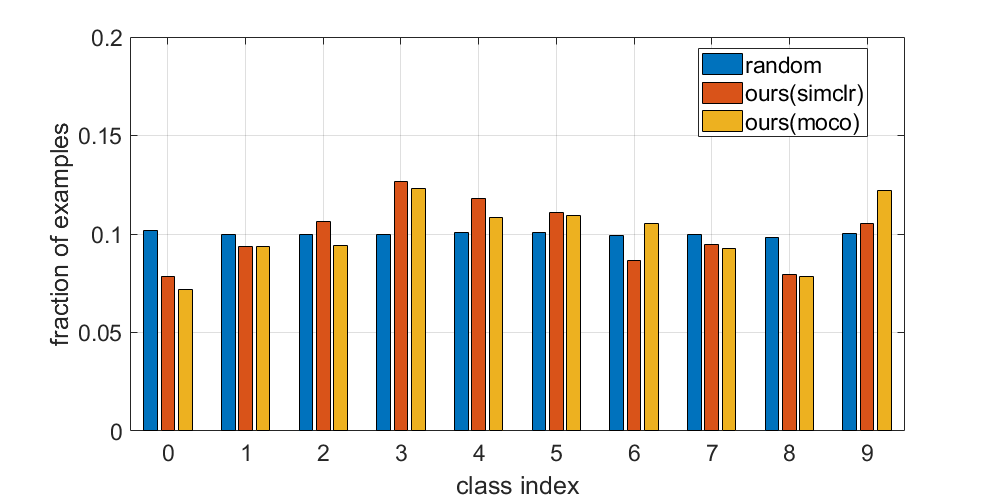}
            \caption{}
        \end{subfigure}
        \begin{subfigure}{55mm}
            \includegraphics[width=55mm]{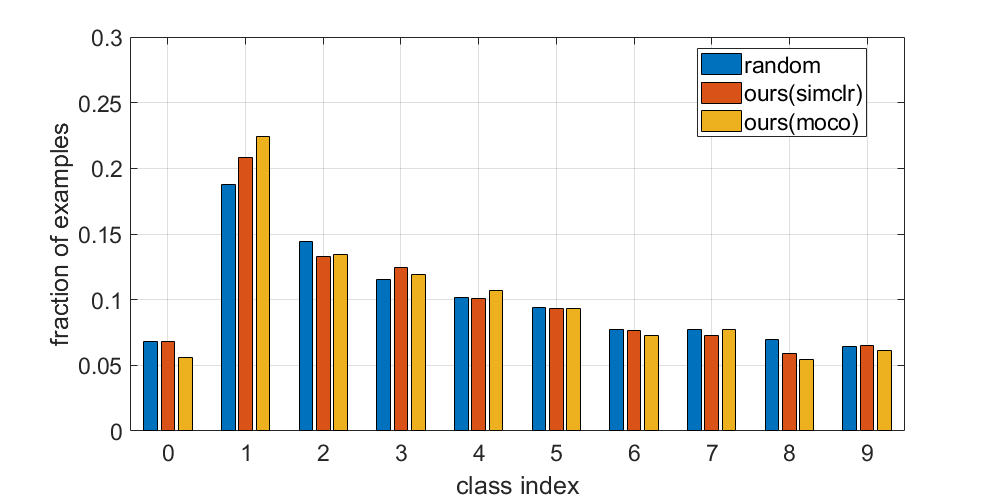}
            \caption{}
        \end{subfigure}
        \begin{subfigure}{55mm}
            \includegraphics[width=55mm]{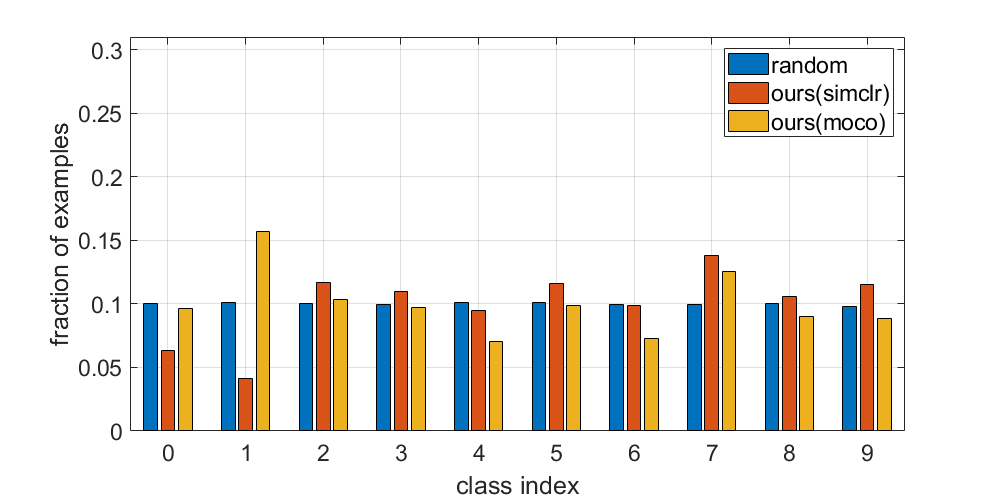}
            \caption{}
        \end{subfigure}
    \end{tabular}
    \caption{\textbf{Distribution of the fraction of examples for each class.} Our coreset exhibits a class imbalance for all the datasets. However, a performance degradation was not observed for subsets above 30\% in size. In addition, for large subsets, the imbalance seems to diminish. }
    \label{fig:imbalance}
    \vspace{-2mm}
\end{figure*}

\section{Ablation Study}
\label{sec:ablation}
In this section, we provide a detailed description of the experiment we conducted to describe our coreset.
First, we attempted to verify that our coreset produces consistent examples that are agnostic to random seeds.
In addition, by varying the model for contrastive learning and target-task learning, we confirmed that our coreset delivers consistent performance agnostic to the model. 
Second, we visually inspected each example to examine the appearance of an image belonging to the coreset and non-coreset.
Finally, because class imbalance inevitably occurs owing to the absence of labels, we plotted the fraction of examples for each class according to the size of the subset. 

\begin{figure*}[t!]
    \begin{tabular}{cccccccccccc}
        \hspace{-4mm}
        \begin{minipage}{3mm}\rotatebox{90}{coreset}\end{minipage}
        \begin{minipage}{15mm}\includegraphics[width=15mm]{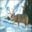}\end{minipage}
        \begin{minipage}{15mm}\includegraphics[width=15mm]{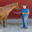}\end{minipage}
        \begin{minipage}{15mm}\includegraphics[width=15mm]{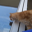}\end{minipage}
        \begin{minipage}{15mm}\includegraphics[width=15mm]{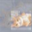}\end{minipage}
        \begin{minipage}{15mm}\includegraphics[width=15mm]{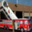}\end{minipage}
        \begin{minipage}{15mm}\includegraphics[width=15mm]{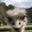}\end{minipage}
        \begin{minipage}{15mm}\includegraphics[width=15mm]{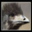}\end{minipage}
        \begin{minipage}{15mm}\includegraphics[width=15mm]{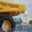}\end{minipage}
        \begin{minipage}{15mm}\includegraphics[width=15mm]{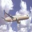}\end{minipage}
        \begin{minipage}{15mm}\includegraphics[width=15mm]{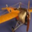}\end{minipage}
        \begin{minipage}{15mm}\includegraphics[width=15mm]{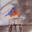}\end{minipage}
        \\
        \hspace{-3mm}
        \begin{minipage}{3mm}\rotatebox{90}{ } \end{minipage}
        \begin{minipage}{15mm}\centering {deer} \vspace{1mm}\end{minipage}
        \begin{minipage}{15mm}\centering {horse} \end{minipage}
        \begin{minipage}{15mm}\centering {cat} \end{minipage}
        \begin{minipage}{15mm}\centering {cat} \end{minipage}
        \begin{minipage}{15mm}\centering {truck} \end{minipage}
        \begin{minipage}{15mm}\centering {bird} \end{minipage}
        \begin{minipage}{15mm}\centering {bird} \end{minipage}
        \begin{minipage}{15mm}\centering {truck} \end{minipage}
        \begin{minipage}{15mm}\centering {airplane} \end{minipage}
        \begin{minipage}{15mm}\centering {airplane} \end{minipage}
        \begin{minipage}{15mm}\centering {bird} \end{minipage}
        \vspace{2mm}
        \\
        \hspace{-4mm}
        \begin{minipage}{3mm}\rotatebox{90}{non-coreset}\end{minipage}
        \begin{minipage}{15mm}\includegraphics[width=15mm]{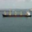}\end{minipage}
        \begin{minipage}{15mm}\includegraphics[width=15mm]{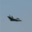}\end{minipage}
        \begin{minipage}{15mm}\includegraphics[width=15mm]{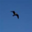}\end{minipage}
        \begin{minipage}{15mm}\includegraphics[width=15mm]{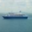}\end{minipage}
        \begin{minipage}{15mm}\includegraphics[width=15mm]{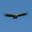}\end{minipage}
        \begin{minipage}{15mm}\includegraphics[width=15mm]{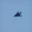}\end{minipage}
        \begin{minipage}{15mm}\includegraphics[width=15mm]{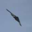}\end{minipage}
        \begin{minipage}{15mm}\includegraphics[width=15mm]{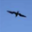}\end{minipage}
        \begin{minipage}{15mm}\includegraphics[width=15mm]{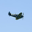}\end{minipage}
        \begin{minipage}{15mm}\includegraphics[width=15mm]{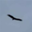}\end{minipage}
        \begin{minipage}{15mm}\includegraphics[width=15mm]{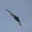}\end{minipage}
        \\
        \hspace{-3mm}
        \begin{minipage}{3mm}\rotatebox{90}{ } \end{minipage}
        \begin{minipage}{15mm}\centering {ship} \end{minipage}
        \begin{minipage}{15mm}\centering {airplane} \end{minipage}
        \begin{minipage}{15mm}\centering {bird} \end{minipage}
        \begin{minipage}{15mm}\centering {ship} \end{minipage}
        \begin{minipage}{15mm}\centering {bird} \end{minipage}
        \begin{minipage}{15mm}\centering {airplane} \end{minipage}
        \begin{minipage}{15mm}\centering {airplane} \end{minipage}
        \begin{minipage}{15mm}\centering {bird} \end{minipage}
        \begin{minipage}{15mm}\centering {airplane} \end{minipage}
        \begin{minipage}{15mm}\centering {bird} \end{minipage}
        \begin{minipage}{15mm}\centering {airplane} \end{minipage}
        \vspace{2mm}
        \\
        \hspace{-4mm}
        \begin{minipage}{3mm}\rotatebox{90}{coreset}\end{minipage}
        \begin{minipage}{15mm}\includegraphics[width=15mm]{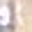}\end{minipage}
        \begin{minipage}{15mm}\includegraphics[width=15mm]{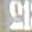}\end{minipage}
        \begin{minipage}{15mm}\includegraphics[width=15mm]{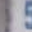}\end{minipage}
        \begin{minipage}{15mm}\includegraphics[width=15mm]{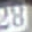}\end{minipage}
        \begin{minipage}{15mm}\includegraphics[width=15mm]{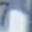}\end{minipage}
        \begin{minipage}{15mm}\includegraphics[width=15mm]{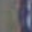}\end{minipage}
        \begin{minipage}{15mm}\includegraphics[width=15mm]{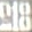}\end{minipage}
        \begin{minipage}{15mm}\includegraphics[width=15mm]{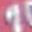}\end{minipage}
        \begin{minipage}{15mm}\includegraphics[width=15mm]{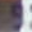}\end{minipage}
        \begin{minipage}{15mm}\includegraphics[width=15mm]{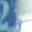}\end{minipage}
        \begin{minipage}{15mm}\includegraphics[width=15mm]{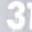}\end{minipage}
        \\
        \hspace{-3mm}
        \begin{minipage}{3mm}\rotatebox{90}{ } \end{minipage}
        \begin{minipage}{15mm}\centering {nine} \end{minipage}
        \begin{minipage}{15mm}\centering {two} \end{minipage}
        \begin{minipage}{15mm}\centering {six} \end{minipage}
        \begin{minipage}{15mm}\centering {eight} \end{minipage}
        \begin{minipage}{15mm}\centering {six} \end{minipage}
        \begin{minipage}{15mm}\centering {nine} \end{minipage}
        \begin{minipage}{15mm}\centering {one} \end{minipage}
        \begin{minipage}{15mm}\centering {seven} \end{minipage}
        \begin{minipage}{15mm}\centering {nine} \end{minipage}
        \begin{minipage}{15mm}\centering {three} \end{minipage}
        \begin{minipage}{15mm}\centering {three} \end{minipage}
        \vspace{2mm}
        \\
        \hspace{-4mm}
        \begin{minipage}{3mm}\rotatebox{90}{non-coreset}\end{minipage}
        \begin{minipage}{15mm}\includegraphics[width=15mm]{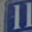}\end{minipage}
        \begin{minipage}{15mm}\includegraphics[width=15mm]{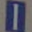}\end{minipage}
        \begin{minipage}{15mm}\includegraphics[width=15mm]{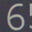}\end{minipage}
        \begin{minipage}{15mm}\includegraphics[width=15mm]{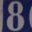}\end{minipage}
        \begin{minipage}{15mm}\includegraphics[width=15mm]{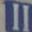}\end{minipage}
        \begin{minipage}{15mm}\includegraphics[width=15mm]{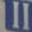}\end{minipage}
        \begin{minipage}{15mm}\includegraphics[width=15mm]{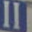}\end{minipage}
        \begin{minipage}{15mm}\includegraphics[width=15mm]{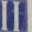}\end{minipage}
        \begin{minipage}{15mm}\includegraphics[width=15mm]{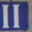}\end{minipage}
        \begin{minipage}{15mm}\includegraphics[width=15mm]{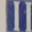}\end{minipage}
        \begin{minipage}{15mm}\includegraphics[width=15mm]{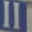}\end{minipage}
        \\
        \hspace{-3mm}
        \begin{minipage}{3mm}\rotatebox{90}{ } \end{minipage}
        \begin{minipage}{15mm}\centering {one} \end{minipage}
        \begin{minipage}{15mm}\centering {one} \end{minipage}
        \begin{minipage}{15mm}\centering {six} \end{minipage}
        \begin{minipage}{15mm}\centering {eight} \end{minipage}
        \begin{minipage}{15mm}\centering {one} \end{minipage}
        \begin{minipage}{15mm}\centering {one} \end{minipage}
        \begin{minipage}{15mm}\centering {one} \end{minipage}
        \begin{minipage}{15mm}\centering {one} \end{minipage}
        \begin{minipage}{15mm}\centering {one} \end{minipage}
        \begin{minipage}{15mm}\centering {one} \end{minipage}
        \begin{minipage}{15mm}\centering {one} \end{minipage}
        \vspace{2mm}
        \\
        \hspace{-4mm}
        \begin{minipage}{3mm}\rotatebox{90}{coreset}\end{minipage}
        \begin{minipage}{15mm}\includegraphics[width=15mm]{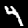}\end{minipage}
        \begin{minipage}{15mm}\includegraphics[width=15mm]{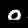}\end{minipage}
        \begin{minipage}{15mm}\includegraphics[width=15mm]{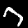}\end{minipage}
        \begin{minipage}{15mm}\includegraphics[width=15mm]{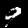}\end{minipage}
        \begin{minipage}{15mm}\includegraphics[width=15mm]{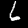}\end{minipage}
        \begin{minipage}{15mm}\includegraphics[width=15mm]{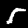}\end{minipage}
        \begin{minipage}{15mm}\includegraphics[width=15mm]{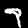}\end{minipage}
        \begin{minipage}{15mm}\includegraphics[width=15mm]{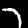}\end{minipage}
        \begin{minipage}{15mm}\includegraphics[width=15mm]{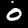}\end{minipage}
        \begin{minipage}{15mm}\includegraphics[width=15mm]{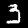}\end{minipage}
        \begin{minipage}{15mm}\includegraphics[width=15mm]{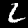}\end{minipage}
        \\
        \hspace{-3mm}
        \begin{minipage}{3mm}\rotatebox{90}{ } \end{minipage}
        \begin{minipage}{15mm}\centering {four} \end{minipage}
        \begin{minipage}{15mm}\centering {zero} \end{minipage}
        \begin{minipage}{15mm}\centering {seven} \end{minipage}
        \begin{minipage}{15mm}\centering {nine} \end{minipage}
        \begin{minipage}{15mm}\centering {six} \end{minipage}
        \begin{minipage}{15mm}\centering {five} \end{minipage}
        \begin{minipage}{15mm}\centering {seven} \end{minipage}
        \begin{minipage}{15mm}\centering {seven} \end{minipage}
        \begin{minipage}{15mm}\centering {zero} \end{minipage}
        \begin{minipage}{15mm}\centering {three} \end{minipage}
        \begin{minipage}{15mm}\centering {two} \end{minipage}
        \vspace{2mm}
        \\
        \hspace{-4mm}
        \begin{minipage}{3mm}\rotatebox{90}{non-coreset}\end{minipage}
        \begin{minipage}{15mm}\includegraphics[width=15mm]{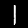}\end{minipage}
        \begin{minipage}{15mm}\includegraphics[width=15mm]{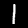}\end{minipage}
        \begin{minipage}{15mm}\includegraphics[width=15mm]{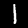}\end{minipage}
        \begin{minipage}{15mm}\includegraphics[width=15mm]{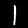}\end{minipage}
        \begin{minipage}{15mm}\includegraphics[width=15mm]{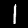}\end{minipage}
        \begin{minipage}{15mm}\includegraphics[width=15mm]{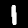}\end{minipage}
        \begin{minipage}{15mm}\includegraphics[width=15mm]{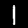}\end{minipage}
        \begin{minipage}{15mm}\includegraphics[width=15mm]{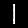}\end{minipage}
        \begin{minipage}{15mm}\includegraphics[width=15mm]{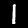}\end{minipage}
        \begin{minipage}{15mm}\includegraphics[width=15mm]{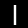}\end{minipage}
        \begin{minipage}{15mm}\includegraphics[width=15mm]{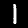}\end{minipage}
        \\
        \hspace{-3mm}
        \begin{minipage}{3mm}\rotatebox{90}{ } \end{minipage}
        \begin{minipage}{15mm}\centering {one} \end{minipage}
        \begin{minipage}{15mm}\centering {one} \end{minipage}
        \begin{minipage}{15mm}\centering {one} \end{minipage}
        \begin{minipage}{15mm}\centering {one} \end{minipage}
        \begin{minipage}{15mm}\centering {one} \end{minipage}
        \begin{minipage}{15mm}\centering {one} \end{minipage}
        \begin{minipage}{15mm}\centering {one} \end{minipage}
        \begin{minipage}{15mm}\centering {one} \end{minipage}
        \begin{minipage}{15mm}\centering {one} \end{minipage}
        \begin{minipage}{15mm}\centering {one} \end{minipage}
        \begin{minipage}{15mm}\centering {one} \end{minipage}
        \\
    \end{tabular}
    \caption{\textbf{Visual inspection for CIFAR10, SVHN, and QMNIST.} Core-set and non-coreset examples for CIFAR10 (\textbf{top two rows}), SVHN (\textbf{middle two rows}) and QMNIST (\textbf{bottom two rows}). Examples belonging to the coreset display a less simple structure than that of the non-coreset. In addition, the majority of non-coreset examples are concentrated in a few classes that could lead to class imbalance.}
    \label{fig:coreset_examples}
\end{figure*}

\begin{table}[ht!]
\caption{\textbf{intersection ratio of five seed runs.} For all datasets, our coreset can be observed to produce consistent examples regardless of the random seed.}
\label{tb:consistent}
\centering
\resizebox{\linewidth}{!}{\begin{tabular}{ccccccc}
\hline\hline
         & Contrastive &  \multicolumn{5}{c}{ Fraction of subset}  \tabularnewline 
 Dataset &  Learning  &  30\%   &   40\%  &   50\%  &   60\%  & 70\% \tabularnewline \hline\hline
         & Random     & 0.01 & 0.02 & 0.06 & 0.13 & 0.24 \tabularnewline 
 CIFAR10 & SimCLR     & 0.90 & 0.92 & 0.93 & 0.95 & 0.96 \tabularnewline 
         & MoCo       & 0.76 & 0.81 & 0.84 & 0.87 & 0.90 \tabularnewline \hline
         & Random     & 0.01 & 0.03 & 0.07 & 0.14 & 0.26 \tabularnewline 
  SVHN   & SimCLR     & 0.91 & 0.92 & 0.93 & 0.94 & 0.96 \tabularnewline 
         & MoCo       & 0.79 & 0.82 & 0.86 & 0.88 & 0.91 \tabularnewline \hline
         & Random     & 0.01  &  0.03 &  0.06 & 0.13 & 0.24 \tabularnewline 
 QMNIST  & SimCLR     & 0.95 & 0.95 & 0.96 & 0.97 & 0.98 \tabularnewline 
         & MoCo       & 0.62 & 0.66 & 0.70 & 0.74 & 0.78 \tabularnewline \hline
\end{tabular}}
\end{table}

\subsection{Consistent Coreset}
\label{subsec:consistent}
As mentioned earlier, an effective coreset algorithm can produce consistent coreset elements across random seeds. This property can be easily verified by calculating the intersection ratio between multiple coresets resulting from multiple executions. These results are presented in Table \ref{tb:consistent}, where each intersection ratio is calculated by dividing the number of intersection elements by the size of the subset. Notably, the intersection ratio of our coreset was significantly greater than that of the random subset across the datasets. Hence, our coreset selects examples that are highly similar despite using multiple random seeds.

\subsection{Model-Agnostic Coreset Selection}
\label{subsec:model-agnostic}
It is apparent that coreset consistency is beneficial in avoiding rerunning coreset selection when the target of the coreset selection model is altered. Hence, we claim that model-agnostic coreset selection is one of the essential properties that any coreset approach has to provide. For our coreset experiment, all the experimental settings were identical to those described in Section \ref{sec:exp}, with the target model, i.e., ResNet101, being the only difference. As depicted in Fig. \ref{fig:agnostic}, our coreset achieved greater performance than that of the random subset. 
For extensive proof, we deployed different architectures, such as wide ResNet (WRN)\cite{zagoruyko2016wide}, ResNeXt\cite{xie2017aggregated}, and DenseNet\cite{huang2017densely}, using a single coreset with SimCLR on CIFAR10. As listed in Fig. \ref{tb:agnostic}, across subset sizes and target model architectures, our coreset consistently yields high performance compared with the random.
\begingroup
\vspace{-2mm}
\setlength{\tabcolsep}{3pt}
\begin{table}[hb!]
\caption{\textbf{Model-agnostic coreset selection task on CIFAR10.} In all the cases, our coreset exhibited high accuracy compared with that of the random subset, regardless of the target model architecture. WRN-n-k ,DenseNet-n, and ResNeXt-n denote WRN with n convolution layers and k widening factor,  DenseNet with n bottleneck layers, and ResNeXt with n layers, respectively.}
\label{tb:agnostic}
\centering
\resizebox{\linewidth}{!}{\begin{tabular}{ccccccc}
\hline\hline
               &  Contrastive  &  \multicolumn{5}{c}{ Fraction of subset }  \tabularnewline 
  Model        &   learning    &    30\%   &    40\%   &    50\%   &    60\%   &  70\% \tabularnewline \hline\hline
 WRN-40-4  &    Random    &   90.94   &   92.14   &   93.17   &   94.04   &  94.30 \tabularnewline 
               &   SimCLR     &    91.82    &  93.23   & 93.61  & 94.46    &  94.94  \tabularnewline \hline
 DenseNet-121   &   Random  &   89.52   &   90.85   &   92.06   &   92.69   &  93.10 \tabularnewline 
               &   SimCLR  &   90.39    &  91.59   & 92.81  & 92.80    &  93.39  \tabularnewline \hline
 ResNeXt-29    &   Random  &   90.68   &   91.99   &   93.68   &   94.30   &  94.61 \tabularnewline 
               &   SimCLR  &   90.71    &  93.13   & 94.01  & 94.54    &  95.05  \tabularnewline \hline
\end{tabular}}
\end{table}
\vspace{-1mm}
\endgroup

\subsection{Visual Inspection}
\label{subsec:visualinspec}
To enable a thorough visual inspection, we identified 11 top/bottom examples (due to page limitations) using a single coreset score with SimCLR, represented in Fig. \ref{fig:coreset_examples}. For simplicity, we referred to the top 11 and bottom 11 as the coreset and non-coreset, respectively. As shown in the figure, the coreset examples exhibit bizarre and peculiar patterns; in contrast, the non-coreset examples primarily comprised simple structures and backgrounds. Furthermore, with respect to class imbalance, the coreset of CIFAR10 predominantly consisted of images of airplanes and birds. Similarly, for both SVHN and QMNIST, most of the top 11 examples consisted of images with a label of 1. It should be noted that our coreset may have been adversely affected by class imbalance. This matter is discussed in the subsequent section.

\subsection{Data Imbalance}
\label{subsec:imbalance}
Unless annotation is provided, it is difficult to prevent interclass imbalance. To examine the class imbalance of our coreset, we displayed the fraction of resulting examples for each class according to the size of the subset, 30\%, 50\%, and 70\%, as depicted in Fig. \ref{fig:imbalance}. A clear imbalance can be observed in our coreset. However, for a subset size above 30\%, the imbalance problem did not seem to affect the classification accuracy. Moreover, as the size of the subset increased, the imbalance tended to decrease. Notably, SVHN training data are inherently unbalanced; hence, even a random subset did not have a uniform distribution. 
Moreover, although our coreset presents a data imbalance, we were able to achieve high accuracy. These facts imply that our coreset search examples contain highly informative content.

\section{Discussion}
\label{sec:discussion}
\begin{figure*}[ht!]
    \centering
    \begin{tabular}{ccc}
    \begin{subfigure}{55mm}\centering{~~~~CIFAR10} \end{subfigure}
    \begin{subfigure}{55mm}\centering{~~~~SVHN} \end{subfigure}
    \begin{subfigure}{55mm}\centering{~~~~QMNIST} \end{subfigure}
    \\
    \begin{subfigure}{55mm}
        \includegraphics[width=55mm]{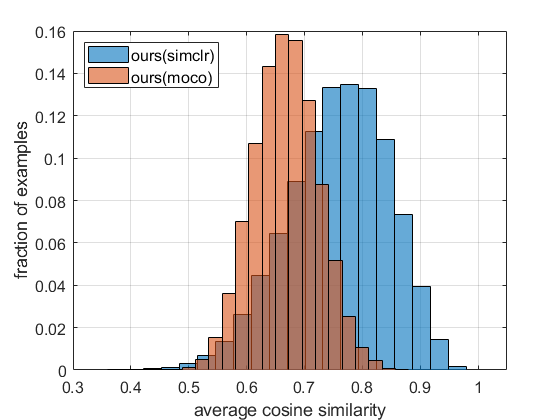}
        \caption{\centering SimCLR : (0.7586, 0.7643) \newline MoCo : (0.6679, 0.6677)}
    \end{subfigure}
    \begin{subfigure}{55mm}
        \includegraphics[width=55mm]{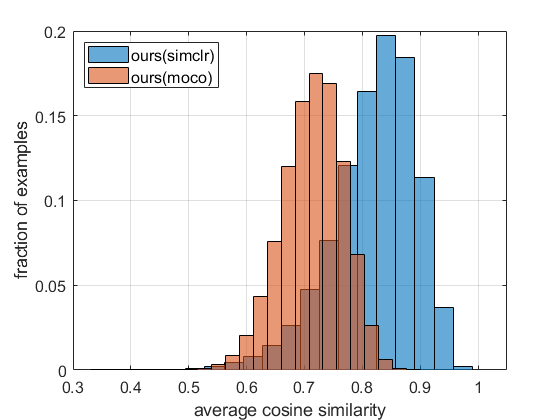}
        \caption{\centering SimCLR : (0.8204, 0.8311) \newline MoCo : (0.7145, 0.7175)}
    \end{subfigure}
    \begin{subfigure}{55mm}
        \includegraphics[width=55mm]{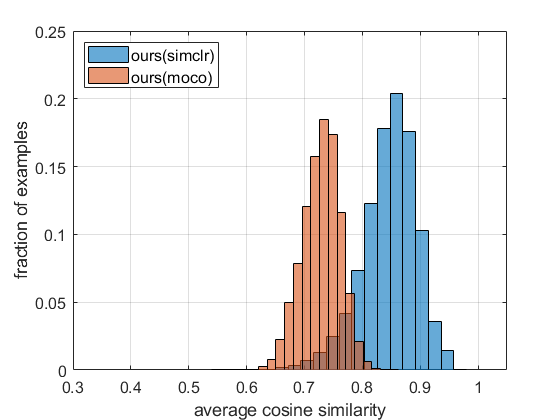}
        \caption{\centering SimCLR : (0.8466, 0.8517) \newline MoCo : (0.7288, 0.7311)}
    \end{subfigure}
    
    \end{tabular}
    \vspace{1mm}
        \caption{\textbf{Distribution of average cossim.} These figures depict the distribution of the average cossim resulting from SimCLR and MoCo for three datasets. $(\cdot, \cdot)$ are a pair of mean and median values. 
        The modes and means of distribution differ depending on the dataset. The mode and mean of QMNIST (the easiest case) are higher than those of CIFAR10 and SVHN.}
    \label{fig:histofcos}
\end{figure*}

\noindent\textbf{Extending the contrastive learning--based score for the core-dataset.} We claimed that the coreset score adequately represents examples that are more informative. To extend this result, we attempted to answer the question of whether it would be possible to distinguish more informative datasets from the others In other words, our claim encouraged us to explore the possibility of employing the contrastive learning--based score to represent the \textbf{core-dataset}. To validate our claim, we plotted the distribution of average cossim for each dataset, as can be seen in Fig.\ref{fig:histofcos}, 
It is evident that relatively more informative datasets are in the order of CIFAR10, SVHN, and QMNIST because these datasets are composed of RGB + objects, RGB + digits, and gray scale + digits, respectively.
Accordingly, the mean and median values for the average cossim values were in the same order. This could imply that the cossim distribution represents the level of difficulty for a dataset; for example, QMNIST is a fairly easy dataset for performing classification compared with CIFAR10 and SVHN; hence, the mean and median of average cossim distribution yielded a large value. We conjecture that our coreset score can possibly be used to identify the core-datasets, and could be the subject of further study in the future.

\section{Conclusion}
\label{sec:conclusion}
In this study, we demonstrated that self-supervised contrastive learning can create a coreset in the absence of a label. Because cossim is readily available from contrastive learning, we identified a metric that effectively measures the coreset score.
Construction of the coreset score based on the average cossim value enabled us to successfully select the coreset from a pool of unlabeled data. This increased the classification accuracy over that of a randomly chosen subset. In addition, our approach yielded results that were comparable to those of methods using supervised learning for coreset selection. 
Although cossim is appropriate for identifying the coreset \textit{per se}, the problem of establishing a new metric still remains. Therefore, in future research, we hope to establish a metric better tailored to coresets. 
Furthermore, when a pool of unlabeled data is gathered from the website and annotation cost should be reduced, 
our study possibly provides an guidance which example to annotate first without sacrificing performance.

{\small
\bibliographystyle{ieee_fullname}
\bibliography{egbib}
}

\end{document}